\documentclass[10pt,twocolumn,letterpaper]{article}

\usepackage{cvpr}
\usepackage{times}
\usepackage{epsfig}
\usepackage{graphicx}
\usepackage{amsmath}
\usepackage{amssymb}
\usepackage{subfigure}
\usepackage{algorithm,algpseudocode,bm}
\usepackage{xcolor}

\usepackage[pagebackref=true,breaklinks=true,letterpaper=true,colorlinks,citecolor=blue,linkcolor=blue,bookmarks=false]{hyperref}

\newcommand{\BG}[1]{\textcolor{blue}{BG: #1}}

\newcommand\cut[1]{}

\cvprfinalcopy 

\def\cvprPaperID{8760} 

\ifcvprfinal\fi
\begin{document}

\title{Rethinking Class-Balanced Methods for Long-Tailed Visual Recognition \\from a Domain Adaptation Perspective}

\author{
Muhammad Abdullah Jamal\textsuperscript{1}\footnotemark \space\space\space
Matthew Brown\textsuperscript{3} \space
Ming-Hsuan Yang\textsuperscript{2,3} \space
Liqiang Wang\textsuperscript{1} \space
Boqing Gong\textsuperscript{3}\vspace{.3em}\\
\textsuperscript{1}University of Central Florida \quad \textsuperscript{2}University of California at Merced\quad \textsuperscript{3}Google
\vspace{-.5em}
}
\maketitle
 \renewcommand*{\thefootnote}{\fnsymbol{footnote}}
 \setcounter{footnote}{1}
 \footnotetext{Work done while M.\ Jamal was an intern at Google.}
 \renewcommand*{\thefootnote}{\arabic{footnote}}
 \setcounter{footnote}{0}






\maketitle

\begin{abstract}
Object frequency in the real world often follows a power law, leading to a mismatch between datasets with long-tailed class distributions seen by a machine learning model and our expectation of the model to perform well on all classes. We analyze this mismatch from a domain adaptation point of view. First of all, we connect existing class-balanced methods for long-tailed classification to target shift, a well-studied scenario in domain adaptation. The connection reveals that these methods implicitly assume that the training data and test data share the same class-conditioned distribution, which does not hold in general and especially for the tail classes. While a head class could contain abundant and diverse training examples that well represent the expected data at inference time, the tail classes are often short of representative training data. To this end, we propose to augment the classic class-balanced learning by explicitly estimating the differences between the class-conditioned distributions with a meta-learning approach. We validate our approach with six benchmark datasets and three loss functions.
\end{abstract}

\section{Introduction} \label{sec-intro}
Big curated datasets, deep learning, and unprecedented computing power are often referred to as the three pillars of recent advances in visual recognition~\cite{krizhevsky2012imagenet,ren2015faster,long2015fully}. As we continue to build the big-dataset pillar, however, the power law emerges as an inevitable challenge. Object frequency in the real world often exhibits a long-tailed distribution where a small number of classes dominate, such as plants and animals~\cite{inaturalist2017,inaturalist}, landmarks around the globe ~\cite{noh2017large}, and common and uncommon objects in contexts~\cite{lin2014microsoft,gupta2019lvis}. 

\begin{figure}
    \centering
    \includegraphics[width=0.9\columnwidth]{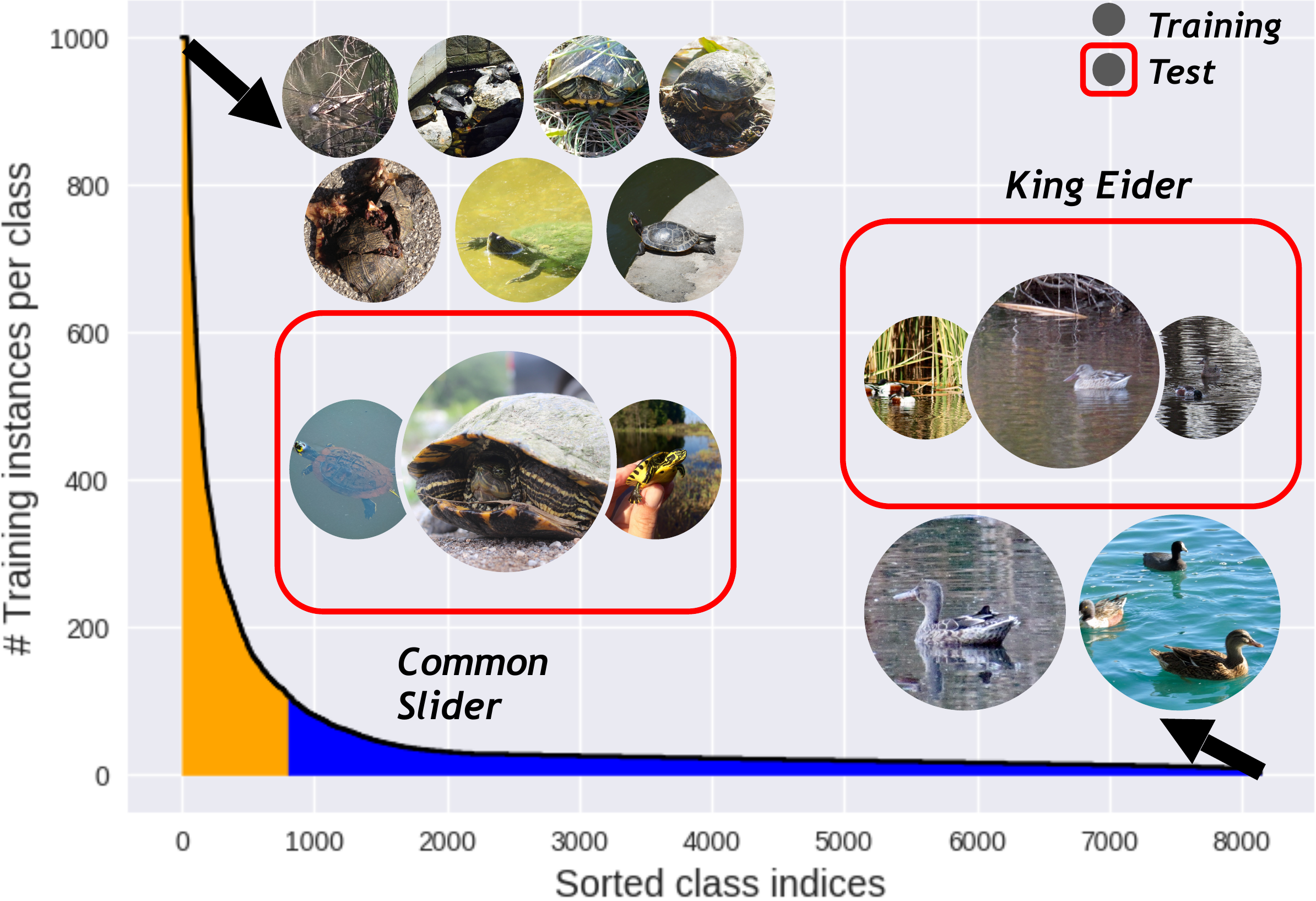}
    \caption{The training set of iNaturalist 2018 exhibits a long-tailed class distribution~\cite{inaturalist}. We connect domain adaptation with the mismatch between the long-tailed training set and our expectation of the trained classifier to perform equally well in all classes. We also view the prevalent class-balanced methods in long-tailed classification as the target shift in domain adaptation, i.e., $P_s(y)\neq P_t(y)$ and \textcolor{blue}{$P_s(x|y)=P_t(x|y)$}, where $P_s$ and $P_t$ are respectively the distributions of the source domain and the target domain, and $x$ and $y$ respectively stand for the input and output of a classifier. We contend that the second part of the target shift assumption does not hold for tail classes, e.g., \textcolor{blue}{$P_s(x|\textit{King Eider})\neq P_t(x|\textit{King Eider})$}, because the limited training images of \textit{King Eider} cannot well represent the data at inference time.}
    \label{fig:longtail}
    \vspace{-10pt}
\end{figure}

In this paper, we propose to investigate long-tailed visual recognition from a domain adaptation point of view. The long-tail challenge is essentially a mismatch problem between datasets with long-tailed class distributions seen by a machine learning model and our expectation of the model to perform well on all classes (and not bias toward the head classes). Conventional visual recognition methods, for instance, training neural networks by a cross-entropy loss, overly fit the dominant classes and fail in the underrepresented tail classes as they implicitly assume that the test sets are drawn i.i.d.\ from the same underlying distribution as the long-tailed training set. Domain adaptation explicitly breaks the assumption~\cite{shimodaira2000improving,saenko2010adapting,gong2012geodesic}. It discloses the inference-time data or distribution (\textit{target domain}) to the machine learning models when they learn from the training data (\textit{source domain}). 

Denote by $P_s(x,y)$ and $P_t(x,y)$ the distributions of a source domain and a target domain, respectively, where $x$ and $y$ are respectively an instance and its class label. In long-tailed visual recognition, the marginal class distribution $P_s(y)$ of the source domain is long-tailed, and yet the class distribution $P_t(y)$ of the target domain is more balanced, e.g., a uniform distribution.

In generic domain adaptation, there could be multiple causes of mismatches between two domains. Covariate shift~\cite{shimodaira2000improving} causes domain discrepancy on the marginal distribution of input, i.e., $P_s(x)\neq P_t(x)$, but often maintains the same predictive function across the domains, i.e., $P_s(y|x)=P_t(y|x)$. Under the target-shift cause~\cite{zhang2013domain}, the domains differ only by the class distributions, i.e., $P_s(y)\neq P_t(y)$ and $P_s(x|y)=P_t(x|y)$, partially explaining the rationale of designing class-balanced weights to tackle the long-tail challenge~\cite{CBLoss,Weakly_supervised,Elkan_csl,zhou2010multi,Mikolov-class_freq,Cui-Finetune,Smote,Undersampling_Drummond,malisiewicz-iccv11,felzenszwalb2009object}. 

These class-balanced methods enable the tail classes to play a bigger role than their sizes suggest in determining the model's decision boundaries. The class-wise weights are inversely related to the class sizes~\cite{Huang_inverse_class_freq,Mikolov-class_freq,Weakly_supervised}. Alternatively, one can derive these weights from the cost of misclassifying an example of one class to another~\cite{Elkan_csl,zhou2010multi}. Cui et al.\ proposed an interesting weighting scheme by counting the ``effective number'' of examples per class~\cite{CBLoss}. Finally,  over/under-sampling head/tail classes~\cite{Cui-Finetune,malisiewicz-iccv11,Smote,Undersampling_Drummond,felzenszwalb2009object} effectively belongs to the same family as the class-balanced weights, although they lead to practically different training algorithms. Section~\ref{sRelated} reviews other methods for coping with the long-tail challenge. 

One the one hand, the plethora of works reviewed above indicate that the target shift, i.e., $P_s(y)\neq P_t(y)$ and $P_s(x|y)=P_t(x|y)$, is generally a reasonable assumption based on which one can design effective algorithms for learning unbiased models from a training set with a long-tailed class distribution. On the other hand, however, our intuition challenges the second part of the target shift assumption; in other words, $P_s(x|y)=P_t(x|y)$ may not hold. While a head class (e.g., \textit{Dog}) of the training set could contain abundant and diverse examples that well represent the expected data at inference time, the tail classes (e.g., \textit{King Eider}) are often short of representative training examples. As a result, training examples drawn from the conditional distribution $P_s(x|\textit{Dog})$ of the source domain can probably well approximate the conditional distribution $P_t(x|\textit{Dog})$ of the target domain, but the discrepancy between the conditional distributions $P_s(x|\textit{King Eider})$ and $P_t(x|\textit{King Eider})$ of the two domains is likely big because it is hard to collect training examples for \textit{King Eider} (cf.\ Figure~\ref{fig:longtail}). 

To this end, we propose to augment the class-balanced learning by relaxing the assumption that the source and target domains share the same conditional distributions $P_s(x|y)$ and $P_t(x|y)$. By explicitly accounting for the differences between them, we arrive at a two-component weight for each training example. The first part is inherited from the classic class-wise weighting, carrying on its effectiveness in various applications. The second part corresponds to the conditional distributions, and we estimate it by the meta-learning framework of learning to re-weight examples~\cite{L2RW}. We make two critical improvements over this framework. One is that we can initialize the weights close to the optima because we have substantial prior knowledge about the two-component weights as a result of our analysis of the long-tailed problem. The other is that we remove two constraints from the framework such that the search space is big enough to cover the optima with a bigger chance.

We conduct extensive experiments on several  datasets, including both long-tailed CIFAR~\cite{Krizhevsky-CIFAR}, ImageNet~\cite{ImageNet}, and Places-2~\cite{Places2}, which are artificially made long-tailed~\cite{CBLoss,OLTR}, and iNaturalist 2017 and 2018~\cite{inaturalist2017,inaturalist}, which are long-tailed by nature. We test  our approach with three different losses (cross-entropy, focal loss~\cite{Focalloss}, and a label-distribution-aware margin loss~\cite{LDAM}). Results validate that our two-component weighting is advantageous over the class-balanced methods.

\section{Related work}\label{sRelated}
Our work is closely related to the class-balanced methods briefly reviewed in Section~\ref{sec-intro}. In this section, we  discuss domain adaptation and the works of other types for tackling the long-tailed visual recognition.

\vspace{-10pt}
\paragraph{Metric learning, hinge loss, and head-to-tail knowledge transfer.} Hinge loss and metric learning are flexible tools for one to handle the long-tailed problem~\cite{LDAM,Huang_inverse_class_freq,Subspace-Cluster,Rangeloss,Hayat-maxmargin,weinberger2006distance}. They mostly contain two major steps. One is to sample or group the data being aware of the long-tailed property, and the other is to construct large-margin losses. Our approach is loss-agnostic, and we show it can benefit different loss functions in the experiments. Another line of research is to transfer knowledge from the head classes to the tail. Yin et al.\ transfer intra-class variance from the head to tail~\cite{Yin-feature-transfer}, Liu et al.\ add a memory module to the neural networks to transfer semantic features~\cite{OLTR}, and Wang et al.\ employ a meta network to regress network weights between different classes~\cite{Modeltail-Wang}.

\vspace{-10pt}
\paragraph{Hard example mining and weighting.}
Hard example mining is prevalent and effective in object detection~\cite{felzenszwalb2009object,malisiewicz-iccv11,ren2015faster,Focalloss}. While it is not particularly designed for the long-tailed recognition, it can indirectly shift the model's focus to the tail classes, from which the hard examples usually originate (cf.~\cite{CBLoss,Dong:hardmining,Freund:boosting} and our experiments). Nonetheless, such methods could be sensitive to outliers or unnecessarily allow a minority of examples to dominate the training. The recently proposed instance weighting by meta-learning methods~\cite{L2RW,MWN} alleviate those issues. Following the general meta-learning principle~\cite{maml,taml,Metasgd}, they set aside a validation set to guide how to weigh the training examples by gradient descent. Similar schemes are used in learning from noisy data~\cite{jiang2018mentornet,FWL,Wang:bayesian}.

\vspace{-10pt}
\paragraph{Domain adaptation.} In real-world applications, there often exist mismatches between the distributions of training data and test data for various reasons~\cite{torralba2011unbiased,gan2016learning,zhang2019curriculum}. Domain adaptation methods aim to mitigate the mismatches so that the learned  models can generalize well to the inference-time data~\cite{shimodaira2000improving,saenko2010adapting,gong2012geodesic,gong2012overcoming}. There are some approaches that handle the imbalance problem in domain adaptation. Zou et al.~\cite{zou2018unsupervised} deal with the class imbalance by controlling the pseudo-label learning and generation using the confidence scores that are normalized class-wise. Yan et al.~\cite{MMD-DA} use a weighted maximum mean discrepancy to handle the class imbalance in unsupervised domain adaptation. We understand the long-tail challenge in visual recognition from the perspective of domain adaptation. While domain adaptation methods need to access a large amount of unlabeled (and sometimes also a small portion of labeled) target domain data, we do not access any inference-time data in our approach. Unlike existing weighting methods in domain adaptation~\cite{cortes2008sample,huang2007correcting,zhang2013domain}, we meta-learn the  weights.

\section{Class balancing as domain adaptation} \label{sec-domain-adaptation}
In this section, we present a detailed analysis of the class-balanced methods~\cite{Huang_inverse_class_freq,Weakly_supervised,CBLoss,Cui-Finetune,malisiewicz-iccv11} for long-tailed visual recognition from the domain adaptation point of view.

Suppose we have a training set (source domain) $\{(x_i,y_i)\}_{i=1}^n$ drawn i.i.d.\ from a long-tailed distribution $P_s(x,y)$ --- more precisely, the marginal distribution $P_s(y)$ of classes are heavy-tailed because, in visual recognition, it is often difficult to collect examples for rare classes. Nonetheless, we expect to learn a visual recognition model to make as few mistakes as possible on all classes:
\begin{align}
\texttt{error} =\mathbb{E}_{P_t(x,y)} L(f(x;\theta),y), \label{eq-error}
\end{align}
where we desire a target domain $P_t(x,y)$ whose marginal class distribution $P_t(y)$ is more balanced (e.g., a uniform distribution) at the inference time, $f(\cdot;\theta)$ is the recognition model parameterized by $\theta$, and $L(\cdot,\cdot)$ is a 0-1 loss. We abuse the notation $L(\cdot,\cdot)$ a little and let it be a differentiable surrogate loss (i.e., cross-entropy) during training.

Next, we apply the importance sampling trick to connect the expected \texttt{error} with the long-tailed source domain,
\begin{align}
&\texttt{error} =\mathbb{E}_{P_t(x,y)} L(f(x;\theta),y) \\
=& \,\mathbb{E}_{P_s(x,y)} L(f(x;\theta),y)P_t(x,y)/P_s(x,y)\\
=&\, \mathbb{E}_{P_s(x,y)} L(f(x;\theta),y) \frac{\textcolor{blue}{\text{$P_t(y)$}}\textcolor{red}{\text{$P_t(x|y)$}}}{\textcolor{blue}{\text{$P_s(y)$}}\textcolor{red}{\text{$P_s(x|y)$}}}\\
:=&\, \mathbb{E}_{P_s(x,y)} L(f(x;\theta),y) \textcolor{blue}{\text{$w_y$}}\textcolor{red}{\text{$(1+\tilde{\epsilon}_{x,y})$}}, \label{eq-importance-sampling}
\end{align}
where $w_y=P_t(y)/P_s(y)$ and $\tilde{\epsilon}_{x,y}=P_t(x|y)/P_s(x|y)-1$. 

Existing class-balanced methods focus on how to determine the class-wise weights $\{w_y\}$ and result in the following objective function for training,
\begin{align}
\min_\theta\quad \frac{1}{n}\sum_{i=1}^n w_{y_i}L(f(x_i;\theta),y_i), \label{eq-class-balanced}
\end{align}
which approximates the expected inference \texttt{error} (eq.~(\ref{eq-importance-sampling})) by assuming $\tilde{\epsilon}_{x,y}=0$ or, in other words, by assuming $P_s(x|y)=P_t(x|y)$ for any class $y$. This assumption is referred to as target shift~\cite{zhang2013domain} in domain adaptation. 

We contend that the assumption of a shared conditional distribution, $P_s(x|y)=P_t(x|y)$, does not hold in general, especially for the tail classes. One may easily compile a representative training set for \textit{Dog}, but not for \textit{King Eider}. We propose to explicitly model the difference $\tilde{\epsilon}_{x,y}$ between the source and target conditional distributions and arrive at an improved algorithm upon the class-balanced methods.

\section{Modeling the conditional differences}\label{sApproach}
For simplicity, we introduce a conditional weight $\epsilon_{x,y}:=w_y\tilde{\epsilon}_{x,y}$ and re-write the expected inference \texttt{error} as 
\begin{align}
&\texttt{error} = \mathbb{E}_{P_s(x,y)}L(f(x;\theta),y)(w_y+\epsilon_{x,y})\\
\approx&\, \frac{1}{n}\sum_{i=1}^n (w_{y_i}+\epsilon_i) L(f(x_i;\theta),y_i), \label{eq-estimated-error}
\end{align}
where the last term is an unbiased estimation of the \texttt{error}. Notably, we do not make the assumption that the conditional distributions of the source and target domains are the same, i.e., we allow $P_s(x|y)\neq P_t(x|y)$ and ${\epsilon}_i\neq 0$. Hence, the weight for each training example consists of two parts. One component is the class-wise weight $w_{y_i}$, and the other is the conditional weight $\epsilon_i$. We need to estimate both components to derive a practical algorithm from eq.~(\ref{eq-estimated-error}) because the underlying distributions of data are unknown --- although we believe the class distribution of the training set must be long-tailed.

\subsection{Estimating the class-wise weights $\{w_{y}\}$} 
We let the class-wise weights resemble the empirically successful design in the literature. In particular, we estimate them by the recently proposed ``effective numbers''~\cite{CBLoss}. Supposing there are $n_y$ training examples for the $y$-th class, we have $w_y\approx(1-\beta)/(1-\beta^{n_y})$ where $\beta\in [0,1)$ is a hyper-parameter with the recommended value $\beta=(n-1)/n$, and $n$ is the number of training examples.

\subsection{Meta-learning the conditional weights $\{\epsilon_i\}$} \label{sec-algorithm}
We estimate the conditional weights by customizing a meta-learning framework~\cite{L2RW}. We describe our approach below and then discuss two critical differences from the original framework in Section~\ref{sec-vs-L2RW}. 

The main idea is to hold out a balanced development set $D$ from the training set and use it to guide the search for the conditional weights that give rise to the best-performing recognition model $f(\cdot;\theta)$ on the development set. Denote by $T$ the remaining training data. We seek the conditional weights $\bm{\epsilon}:=\{\epsilon_i\}$ by solving the following problem, 
\begin{align}
\min_{\bm{\epsilon}} \quad \frac{1}{|D|}\sum_{\textcolor{blue}{i\in D}}L(f(x_i;\theta^*(\bm{\epsilon})),y_i) \text{ with} \label{eq-eval}\\
 \theta^*(\bm{\epsilon}) \leftarrow \arg\min_\theta \frac{1}{|T|}\sum_{\textcolor{red}{i\in T}}(w_{y_i}+\epsilon_i)L(f(x_i;\theta),y_i) \label{eq-estimated-error-T}
\end{align}
where we do \textit{not} weigh the losses over the development set which is already balanced. Essentially, the problem  above searches for the optimal conditional weights  such that, after we learn a recognition model $f(\cdot;\theta)$ by minimizing the  \texttt{error} estimation (eqs~(\ref{eq-estimated-error-T}) and (\ref{eq-estimated-error})), the model performs the best on the development set (eq.~(\ref{eq-eval})).  

It would be daunting to solve the problem above by  brute-force search, e.g., iterating all the possible sets $\{\bm{\epsilon}\}$ of conditional weights. Even if we can, it is computationally prohibitive to train for each set of weights a recognition model $f(\cdot;\theta^*(\bm{\epsilon}))$ and then find out the best model from all. 
 
Instead, we modify the meta-learning framework~\cite{L2RW} and search for the conditional weights in a greedy manner. It interleaves the quest for the weights $\bm{\epsilon}$ with the updates to the model parameters $\theta$, given current time step $t$,
\begin{align}
\tilde{\theta}^{t+1}(\bm{\epsilon}^{t}) &\leftarrow \theta^t-\eta \frac{\partial \sum_{\textcolor{red}{i\in T}}(w_{y_i}+\epsilon_i^t)L(f(x_i;\theta^t),y_i)}{\partial \theta} \notag\\
\bm{\epsilon}^{t+1} &\leftarrow \bm{\epsilon}^t - \tau \frac{\partial \sum_{\textcolor{blue}{i\in D}}L(f(x_i;\tilde{\theta}^{t+1}(\bm{\epsilon}^{t})),y_i)}{\partial \bm{\epsilon}} \notag \\
{\theta}^{t+1} &\leftarrow \theta^t-\eta \frac{\partial \sum_{\textcolor{red}{i\in T}}(w_{y_i}+\epsilon_i^{t+1})L(f(x_i;\theta^t),y_i)}{\partial \theta}. \notag
\end{align}
The first equation tries a one-step gradient descent for $\theta^t$ using the losses weighted by the current conditional weights $\bm{\epsilon}^t$ (plus the class-wise weights). The updated model parameters $\tilde{\theta}^{t+1}(\bm{\epsilon}^t)$ are then scrutinized on the balanced development set $D$, which updates the conditional weights by one step. The updated weights $\bm{\epsilon}^{t+1}$ are better than the old ones, meaning that the model parameters $\theta^{t+1}$ returned by the last equation should give rise to smaller recognition error on the development set than $\tilde{\theta}^{t+1}$ do. Starting from $\theta^{t+1}$ and $\bm{\epsilon}^{t+1}$, we then move on to the next round of updates. We present our overall algorithm in the next section.

\begin{center}
\begin{algorithm}[t]
  \begin{algorithmic}[1]
\Require Training set $T$, balanced development set $D$
\Require Class-wise weights $\{w_y\}$ estimated by using~\cite{CBLoss}
\Require Learning rates $\eta$ and $\tau$, stopping steps $t_1$ and $t_2$
\Require Initial parameters $\theta$ of the recognition network

\For{$t=1, 2, \cdots, t_1$}
   \State Sample a mini-batch $B$ from the training set $T$
   \State Compute loss 
   $\mathcal{L}_{B}=\frac{1}{|B|}\sum_{i\in B}L(f(x_{i};\theta),y_{i})$ 
   \State  Update $\theta\leftarrow\theta - \eta \nabla_\theta \mathcal{L}_{B}$

\EndFor

\For{$t=t_1+1,\cdots,t_1+t_2$}

    \State Sample a mini-batch $B$ from the training set $T$
    
    \State Set $\epsilon_i\leftarrow 0, \forall i\in B$, and denote by $\bm{\epsilon}:=\{\epsilon_i,i\in B\}$
    
    \State Compute 
    $\mathcal{L}_{B}=\frac{1}{|B|}\sum_{i\in B}(w_{y_i}+\epsilon_i)L(f(x_{i};\theta),y_{i})$ 
    
    \State Update \textcolor{red}{$\tilde{\theta}(\bm{\epsilon})$} $\leftarrow\theta - \eta   \nabla_\theta \mathcal{L}_{B}$
   
    \State \textcolor{blue}{Sample $B_{d}$ from the balanced development set $D$}
    
    \State \textcolor{blue}{Compute  
    $\mathcal{L}_{B_{d}}=\frac{1}{|B_{d}|}\sum_{i\in B_{d}} L(f(x_{i};\textcolor{red}{\tilde{\theta}(\bm{\epsilon})}),y_{i})$}
    
    \State \textcolor{blue}{Update $\bm{\epsilon}\leftarrow\bm{\epsilon} - \tau\nabla_\epsilon  \mathcal{L}_{B_d}$}
    
    \State Compute new loss with the updated \textcolor{blue}{$\bm{\epsilon}$}
    
           $\tilde{\mathcal{L}}_{B}=\frac{1}{|B|}\sum_{i\in B}(w_{y_i}+\textcolor{blue}{\epsilon_i}) L(f(x_{i};\theta),y_{i})$
           
    \State  Update $\theta \leftarrow\theta-\eta \nabla_\theta \tilde{\mathcal{L}}_{B}$

\EndFor

\caption{Meta-learning for long-tailed recognition}
\label{algo:reweighting}
 \end{algorithmic}
\end{algorithm} 
\end{center}

\vspace{-15pt}
\subsection{Overall algorithm and discussion} \label{sec-vs-L2RW}
We are ready to present ~\textbf{Algorithm}~\ref{algo:reweighting} for long-tailed visual recognition. The discussions in the previous sections consider all the training examples in a batch setting. \textbf{Algorithm}~\ref{algo:reweighting} customizes it into a stochastic setting so that we can easily integrate it with deep learning frameworks\cut{ --- we use TensorFlow for experiments}. 

There are two learning stages in the algorithm. In the first stage (lines 1--5), we train the neural recognition network $f(\cdot;\theta)$ by using the conventional cross-entropy loss over the long-tailed training set. The second stage (lines 6--16) meta-learns the conditional weights by resorting to a balanced development set and meanwhile continues to update the recognition model. We highlight the part for updating the conditional weights in lines 11--13.

\vspace{-10pt}
\paragraph{Discussion.} It is worth noting some seemingly small and yet fundamental differences between our algorithm and the learning to re-weight (L2RW) method~\cite{L2RW}. Conceptually, while we share the same meta-learning framework as L2RW, both the class-wise weight, $w_y=P_t(y)/P_s(y)$, and the conditional weight,
    $\epsilon_{x,y}=w_y\tilde{\epsilon}_{x,y}={P_t(y)}/{P_s(y)}\big({P_t(x|y)}/{P_s(x|y)}-1\big)$,
have principled interpretations as oppose to a general per-example weight in L2RW. We will explore other machine  learning frameworks (e.g.,~\cite{bickel2009discriminative,sugiyama2007covariate}) to learn the conditional weights in future work, but the interpretations of them remain the same. 

Algorithmically, unlike L2RW, we employ two-component weights, estimate the class-wise components by a different method~\cite{CBLoss}, do \textit{not} clip negative weights $\{\epsilon_i\}$ to 0, and do \emph{not} normalize them such that they sum to 1 within a mini-batch. The clipping and normalization operations in L2RW unexpectedly reduce the search space of the weights, and the normalization is especially troublesome as it depends on the mini-batch size. Hence, if the optimal weights actually lie outside of the reduced search space, there is no chance to hit the optima by L2RW. In contrast, our algorithm searches for each conditional weight $\epsilon_i$ in the full real space. One may wonder whether or not our total effective weight, $w_{y_i}+\epsilon_i$, could become negative. Careful investigation reveals that it never goes below 0 in our experiments, likely due to that the good initialization (as explained below) to the conditional weights makes it unnecessary to update the weights too wildly by line 13 in \textbf{Algorithm}~\ref{algo:reweighting}.

Computationally, we provide proper initialization to both the conditional weights, by $\epsilon_i\leftarrow 0$ (line 8), and the model parameters $\theta$ of the recognition network, by pre-training the network with a vanilla cross-entropy loss (lines 1--5). As a result, our algorithm is more stable than L2RW (cf.\ Section~\ref{sec-exp-cifar}). Note that 0 is a reasonable \textit{a priori} value for the conditional weights thanks to the promising results obtained by existing class-balanced methods. Those methods assume that the discrepancy is as small as 0 between the conditional distributions of the source and target domains, meaning that $P_t(x|y)/P_s(x|y)-1$ is close to 0, so are the conditional weights $\{\epsilon_i\}$. Hence, our approach should perform at worst the same as the class-balanced method~\cite{CBLoss} by initializing the conditional weights to 0 (and the class-wise weights by~\cite{CBLoss}).  
\section{Experiments}\label{sExps}

\paragraph{Datasets.}
We evaluate and ablate our approach on six datasets of various scales, ranging from the manually created long-tailed CIFAR-10 and CIFAR-100~\cite{CBLoss},  ImageNet-LT, and Places-LT~\cite{OLTR}, to the naturally long-tailed iNaturalist 2017~\cite{inaturalist2017} and 2018~\cite{inaturalist}. Following~\cite{CBLoss}, \cut{Ranking all classes of a dataset by their sizes non-increasingly,} we define the \emph{imbalance factor (IF)} of a dataset as the class size of the first head class divided by the size of the last tail class.

\begin{table*}
\centering
\small
\caption {\label{tab:datasets} Overview of the six datasets used in our experiments. (IF stands for the imbalance factor)}
\vspace{2pt}
\begin{tabular}{l|r|r|r|r|r|r|r}
\hline
Dataset & \# Classes & IF & \# Train.\ img.\ & Tail class size & Head class size  & \# Val.\ img.\  & \# Test img.\\ \hline
CIFAR-LT-10\cut{~\cite{Krizhevsky-CIFAR,CBLoss}} & 10 & 1.0--200.0 & 50,000--11,203 & 500--25  & 5,000 & -- & 10,000\\ \hline
CIFAR-LT-100\cut{~\cite{Krizhevsky-CIFAR,CBLoss}} & 100 & 1.0--200.0 & 50,000--9,502  & 500--2 & 500 & -- & 10,000 \\ \hline
iNat 2017\cut{~\cite{inaturalist2017}} & 5,089 & 435.4 & 579,184 & 9 & 3,919 & 95,986 & -- \\ \hline
iNat 2018\cut{~\cite{inaturalist}} & 8,142 & 500.0 & 437,513 & 2 &  1,000 & 24,426 & -- \\ \hline
ImageNet-LT\cut{~\cite{ImageNet,OLTR}} & 1,000 & 256.0 & 115,846 & 5 & 1,280 & 20,000 & 50,000 \\ \hline
Places-LT\cut{~\cite{Places2,OLTR}} & 365 & 996.0 & 62,500 & 5 & 4,980 & 7,300 & 36,500 \\ \hline
\end{tabular}
\vspace{-10pt}
\end{table*}

\vspace{-5pt}
\begin{description} \setlength\itemsep{-1pt}

\item[Long-Tailed CIFAR (CIFAR-LT):] The original CIFAR-10 (CIFAR-100) dataset contains 50,000 training images and 10,000 test images of size 32x32 uniformly falling into 10 (100) classes~\cite{Krizhevsky-CIFAR}. Cui et al.~\cite{CBLoss} created long-tailed versions by randomly removing training examples. In particular, the number of examples dropped from the $y$-th class is $n_{y} \mu^{y}$, where $n_{y}$ is the original number of training examples in the class and $\mu\in(0, 1)$. By varying $\mu$, we arrive at six training sets, respectively, with the imbalance factors (IFs) of 200, 100, 50, 20, 10, and 1, where IF=$1$ corresponds to the original datasets. We do not change the test sets, which are balanced. We randomly select ten training images per class as our development set $D$.

\item \textbf{ImageNet-LT}: In spirit similar to the long-tailed CIFAR datasets, Liu et al.~\cite{OLTR} introduced a long-tailed version of ImageNet-2012~\cite{ImageNet} called ImageNet-LT. It is created by firstly sampling the class sizes from a Pareto distribution with the power value $\alpha=6$, followed by sampling the corresponding number of images for each class. The resultant dataset has 115.8K training images in 1,000 classes, and its imbalance factor is 1280/5. The authors have also provided a validation set with 20 images per class, from which we sample ten images to construct our development set $D$. The original balanced ImageNet-2012 validation set is used as the test set (50 images per class). 

\item \textbf{Places-LT}: Liu et al.~\cite{OLTR} have also created a Places-LT dataset by sampling from Places-2~\cite{Places2} using the same strategy as above. It contains 62.5K training images from 365 classes with an imbalance factor 4980/5. This large imbalance factor indicates that it is more challenging than ImageNet-LT. Places-LT has 20 (100) validation (test) images per class. Our development set $D$ contains ten images per class randomly selected from the validation set.

\item \textbf{iNaturalist (iNat) 2017 and 2018}: The iNat 2017~\cite{inaturalist2017} and 2018~\cite{inaturalist} are real-world fine-grained visual recognition datasets that naturally exhibit long-tailed class distributions. iNat 2017 (2018) consists of 579,184 (435,713) training images in 5,089 (8,142) classes, and its imbalance factor is 3919/9 (1000/2). We use the official validation sets to test our approach. We select five (two) images per class from the training set of iNat 2017 (2018) for the development set.
\vspace{-5pt}
\end{description}

Table~\ref{tab:datasets} gives an overview of the six datasets used in the following experiments. 

\vspace{-10pt}
\paragraph{Evaluation Metrics.}
As the test sets are all balanced, we simply use the top-$k$ error as the evaluation metric. We report results for $k=1,3,5$.

\subsection{Object recognition with CIFAR-LT} \label{sec-exp-cifar}
We run both comparison experiments and ablation studies with CIFAR-LT-10 and CIFAR-LT-100. We use ResNet-32~\cite{Resnet} in the experiments.

\vspace{-10pt}
\paragraph{Competing methods.}
We compare our approach to the following competing ones.

\vspace{-10pt}
\begin{itemize}   \setlength\itemsep{-3pt}

\item \textbf{Cross-entropy training.} This is the baseline that trains ResNet-32 using the vanilla cross-entropy loss.

\item \textbf{Class-balanced loss~\cite{CBLoss}.}
It weighs the conventional losses by class-wise weights, which are estimated based on effective numbers. We apply this class-balanced weighting to three different losses: cross-entropy, the focal loss~\cite{Focalloss}, and the recently proposed label-distribution-aware margin loss~\cite{LDAM}.

\item \textbf{Focal loss~\cite{Focalloss}.} The focal loss can be understood as a smooth version of hard example mining. It does not directly tackle the long-tailed recognition problem. However, it can penalize the examples of tail classes more than those of the head classes if the network is biased toward the head classes during training.

\item \textbf{Label-distribution-aware margin loss~\cite{LDAM}.} It dynamically tunes the margins between classes according to their degrees of dominance in the training set. 

\item \textbf{Class-balanced fine-tuning~\cite{Cui-Finetune}.} The main idea is to first train the neural network with the whole imbalanced training set and then fine-tune it on a balanced subset of the training set. 

\item \textbf{Learning to re-weight (L2RW)~\cite{L2RW}.} It weighs training examples by meta-learning. Please see Section~\ref{sec-vs-L2RW} for more discussions about L2RW and our approach.

\item \textbf{Meta-weight net~\cite{MWN}.} Similarly to L2RW, it also weighs examples by a meta-learning method except that it regresses the weights by a multilayer perceptron.

\vspace{-10pt}
\end{itemize}

\vspace{-10pt}
\paragraph{Implementation details.}
For the first two baselines, we use the code of \cite{CBLoss} to set the learning rates and other hyperparameters. We train the L2RW model using an initial learning rate of 1e-3. We decay the learning rate by 0.01 at the 160th and 180th epochs. For our approach, we use an initial learning rate of 0.1 and then also decay the learning rate at the 160th and 180th epochs by 0.01. The batch size is 100 for all experiments. We train all models on a single GPU using the stochastic gradient descent with momentum.

\begin{table*}
\centering
\caption {\label{tab:CIFAR-10} Test top-1 errors (\%) of ResNet-32 on CIFAR-LT-10 under different imbalance settings. * indicates results reported in ~\cite{MWN}.}
\vspace{2pt}
\resizebox{0.75\textwidth}{!}{%
\begin{tabular}{l|c|l|l|l|l|l}
\hline

\hline

\hline
Imbalance factor & 200 & \multicolumn{1}{c|}{100} & \multicolumn{1}{c|}{50} & \multicolumn{1}{c|}{20} & \multicolumn{1}{c|}{10} & \multicolumn{1}{c}{1} \\ \hline
Cross-entropy training & 34.32 & 29.64 & 25.19 & 17.77 & 13.61 & 7.53/7.11* \\ \hline
Class-balanced cross-entropy loss~\cite{CBLoss} & 31.11 & 27.63 & 21.95 & 15.64 &13.23 & 7.53/7.11* \\ \hline

\begin{tabular}[c]{@{}l@{}}Class-balanced fine-tuning~\cite{Cui-Finetune}\\ Class-balanced fine-tuning~\cite{Cui-Finetune}*\end{tabular} & \begin{tabular}[c]{@{}c@{}}33.76\\ 33.92\end{tabular} & \begin{tabular}[c]{@{}l@{}}28.66\\ 28.67\end{tabular} & \begin{tabular}[c]{@{}l@{}}22.56\\ 22.58\end{tabular} &
\begin{tabular}[c]{@{}l@{}}16.78\\ 13.73\end{tabular} & \begin{tabular}[c]{@{}l@{}}16.83\\ 13.58\end{tabular} &
\begin{tabular}[c]{@{}l@{}}7.08\\ \textbf{6.77}\end{tabular} \\ \hline

\begin{tabular}[c]{@{}l@{}}L2RW~\cite{L2RW}\\ L2RW~\cite{L2RW}*\end{tabular} & \begin{tabular}[c]{@{}c@{}}33.75\\ 33.49\end{tabular} & \begin{tabular}[c]{@{}l@{}}27.77\\ 25.84\end{tabular} & \begin{tabular}[c]{@{}l@{}}23.55\\ 21.07\end{tabular} &  \begin{tabular}[c]{@{}l@{}}18.65\\ 16.90\end{tabular} & \begin{tabular}[c]{@{}l@{}}17.88\\ 14.81\end{tabular} & \begin{tabular}[c]{@{}l@{}}11.60\\ 10.75\end{tabular} \\ \hline

Meta-weight net~\cite{MWN} & 32.8 & 26.43 & 20.9 & 15.55 & 12.45 & 7.19  \\ \hline


{\bf Ours with cross-entropy loss} & \multicolumn{1}{l|}{\bf 29.34} & {\bf 23.59} & {\bf 19.49} & \textbf{13.54} & \textbf{11.15}  & 7.21 \\ \hline 
 
 
 

\hline 

\hline

Focal loss~\cite{Focalloss} & 34.71  & 29.62 & 23.29
 & 17.24 & 13.34 & 6.97 \\ \hline

Class-balanced focal Loss~\cite{CBLoss} & 31.85  & 25.43 & 20.78 & 16.22 & 12.52 & 6.97 \\ \hline

{\bf Ours with focal Loss} & {\bf 25.57} & {\bf 21.1} & \textbf{17.12}  & {\bf 13.9} & {\bf 11.63} & 7.19 \\ \hline

\hline 

\hline


LDAM loss~\cite{LDAM} (results reported in paper) & - & 26.65 & - & - & 13.04 & 11.37 \\ \hline

LDAM-DRW~\cite{LDAM} (results reported in paper) & - & 22.97 & - & - & \textbf{11.84}  & - \\ \hline

{\bf Ours with LDAM loss} & \textbf{22.77} & \textbf{20.0} & {\bf 17.77}  & {\bf 15.63} & 12.6 & {\bf 10.29} \\ \hline

\end{tabular}
}%
\vspace{-5pt}
\end{table*}



\begin{table*}
\centering
\caption {\label{tab:CIFAR-100} Test top-1 errors (\%) of ResNet-32 on CIFAR-LT-100 under different imbalance settings. * indicates results reported in ~\cite{MWN}.}
\vspace{2pt}

\resizebox{0.75\textwidth}{!}{%
\begin{tabular}{l|c|l|l|l|l|l}
\hline

\hline

\hline
Imbalance factor & 200 & \multicolumn{1}{c|}{100} & \multicolumn{1}{c|}{50} & \multicolumn{1}{c|}{20} & \multicolumn{1}{c|}{10} & \multicolumn{1}{c}{1} \\ \hline
Cross-entropy training & 65.16  & 61.68 & 56.15
 &48.86 &44.29 & 29.50 \\ \hline
Class-balanced cross-entropy loss~\cite{CBLoss} & 64.30 &61.44  & 55.45 &48.47  & 42.88 & 29.50 \\ \hline

\begin{tabular}[c]{@{}l@{}}Class-balanced fine-tuning~\cite{Cui-Finetune}\\ Class-balanced fine-tuning~\cite{Cui-Finetune}*\end{tabular} & 
\begin{tabular}[c]{@{}c@{}}61.34\\61.78\end{tabular} & \begin{tabular}[c]{@{}l@{}}58.5\\58.17 \end{tabular} & \begin{tabular}[c]{@{}l@{}}53.78\\53.60 \end{tabular} &
\begin{tabular}[c]{@{}l@{}}47.70\\47.89 \end{tabular} &
\begin{tabular}[c]{@{}l@{}}42.43\\42.56 \end{tabular} & \begin{tabular}[c]{@{}l@{}}29.37\\29.28 \end{tabular} \\ \hline

\begin{tabular}[c]{@{}l@{}}L2RW~\cite{L2RW}\\ L2RW~\cite{L2RW}*\end{tabular} & \begin{tabular}[c]{@{}c@{}}67.00\\66.62 \end{tabular} & \begin{tabular}[c]{@{}l@{}}61.10\\59.77 \end{tabular} & \begin{tabular}[c]{@{}l@{}}56.83\\55.56 \end{tabular} &
\begin{tabular}[c]{@{}l@{}}49.25\\48.36 \end{tabular}& 
\begin{tabular}[c]{@{}l@{}}47.88\\46.27 \end{tabular}&
\begin{tabular}[c]{@{}l@{}}36.42\\35.89 \end{tabular} \\ \hline

Meta-weight net~\cite{MWN} & 63.38 & 58.39  & 54.34  &46.96  & 41.09 & 29.9 \\ \hline


{\bf Ours with cross-entropy loss} & \multicolumn{1}{l|}{\bf 60.69} & {\bf 56.65} & { \bf 51.47} & \textbf{44.38} & \textbf{40.42} & \textbf{28.14}
\\ \hline

\hline 

\hline

Focal Loss~\cite{Focalloss} & 64.38  & 61.59  & 55.68
 & 48.05 & 44.22 &  \textbf{28.85} \\ \hline

Class-balanced focal Loss~\cite{CBLoss} & 63.77 & 60.40  & 54.79 & 47.41 & 42.01 & \textbf{28.85} \\ \hline

{\bf Ours with focal loss} & \textbf{60.66} & \textbf{55.3} & \textbf{49.92}  & \textbf{44.27} & \textbf{40.41} & 29.15 \\ \hline

\hline

\hline 


LDAM Loss~\cite{LDAM} (results reported in paper) &  - &  60.40  & -  & -  &  43.09 & - \\ \hline

LDAM-DRW~\cite{LDAM} (results reported in paper)& - & 57.96  & - & - & \textbf{41.29}  & - \\ \hline

{\bf Ours with LDAM loss} & {\bf 60.47} & \textbf{55.92} & \textbf{50.84} & \textbf{47.62} & 42.0 & - \\ \hline

\end{tabular}
}%
\vspace{-10pt}
\end{table*}

\vspace{-10pt}
\paragraph{Results.}
Table~\ref{tab:CIFAR-10} shows the classification errors of ResNet-32 on the long-tailed CIFAR-10 with different imbalance factors. We group the competing methods into three sessions according to which basic loss they use (cross-entropy, focal~\cite{Focalloss}, or LDAM~\cite{LDAM}). We test our approach with all three losses. We can see that our method outperforms the competing ones in each session by notable margins. Although the focal loss and the LDAM loss already have the capacity of mitigating the long-tailed issue, respectively, by penalizing hard examples and by distribution-aware margins, our method can further boost their performances. In general, the advantages of our approach over existing ones become more significant as the imbalance factor increases. When the dataset is balanced (the last column), our approach does not hurt the performance of vanilla losses compared to L2RW. We can draw about the same observations as above for the long-tailed CIFAR-100 from Table~\ref{tab:CIFAR-100}.

\begin{figure*}
    \centering
    \subfigure{
    \includegraphics[width=0.3\textwidth]{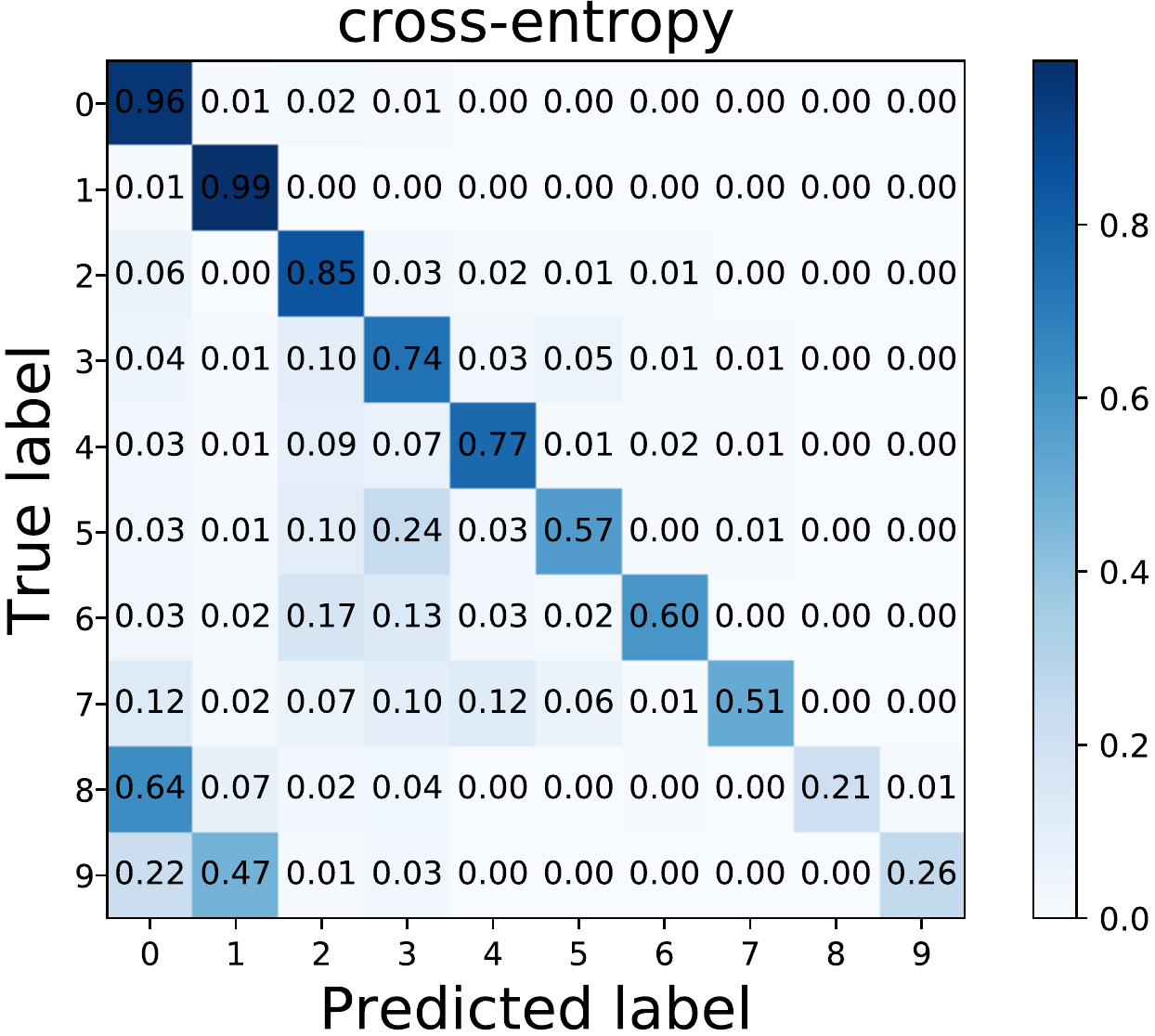}
    }
    \centering
    \subfigure{
    \includegraphics[width=0.3\textwidth]{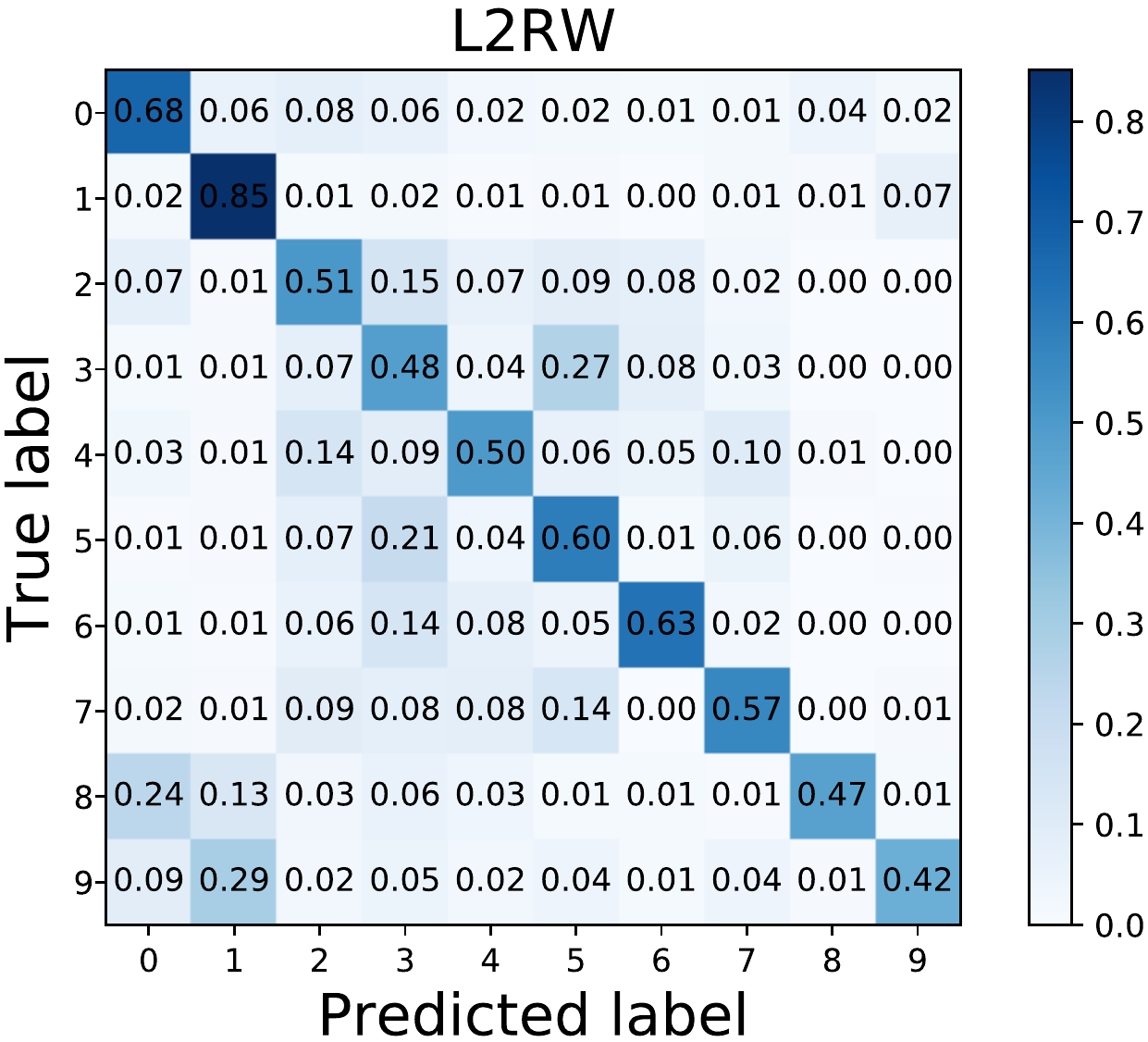}
    }
    \centering
    \subfigure{
    \includegraphics[width=0.3\textwidth]{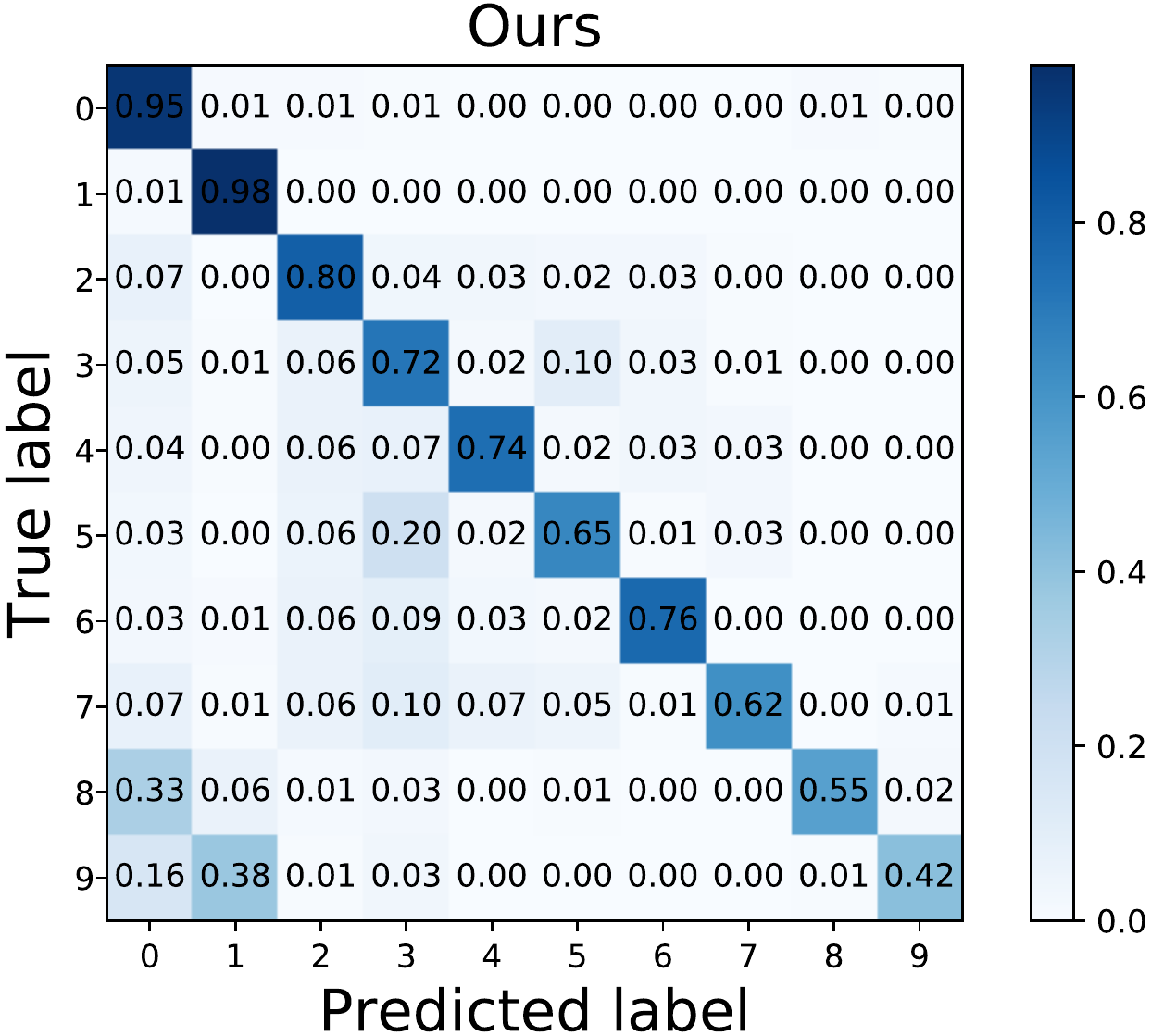}
    }
    \caption{Confusion matrices by the cross-entropy training, L2RW, and our method on CIFAR-LT-10 (the imbalance factor is 200).}
    \label{fig:confusion_matrices}
    \vspace{-5pt}
\end{figure*}

\vspace{-10pt}
\paragraph{Where does our approach work?} Figure~\ref{fig:confusion_matrices} presents three confusion matrices respectively by the models of the cross-entropy training, L2RW, and our method on CIFAR-LT-10. The imbalance factor is 200. Compared with the cross-entropy model, L2RW  improves the accuracies on the tail classes and yet sacrifices the accuracies for the head classes.  In contrast, ours maintains about the same performance as the cross-entropy model on the head classes and meanwhile significantly improves the accuracies for the last five tail classes.

\begin{figure}
   \centering
    \begin{tabular}{ll}
    \includegraphics[width=0.23\textwidth]{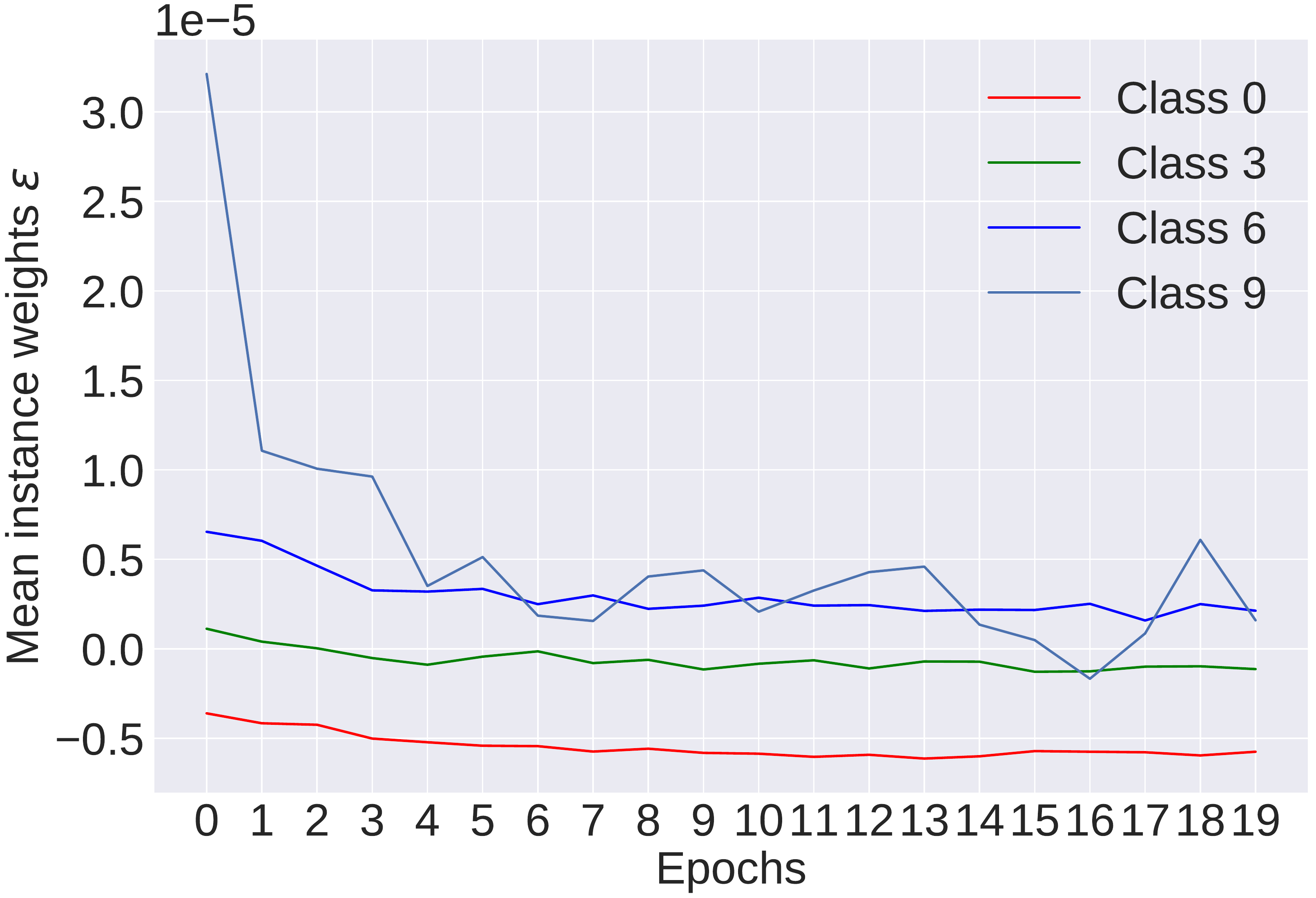}
    &
    \includegraphics[width=0.23\textwidth]{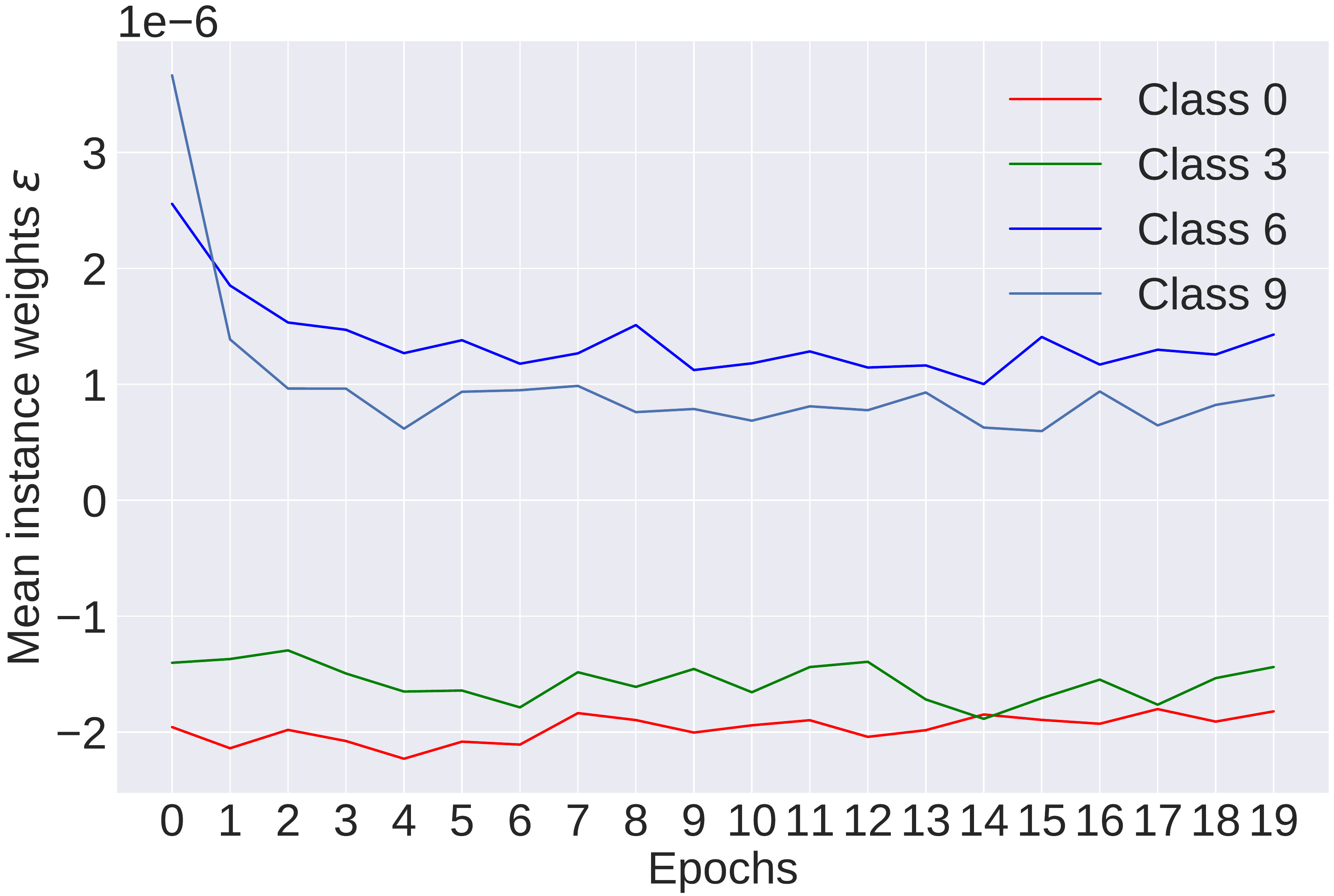}
    \end{tabular}
    \vspace{-7pt}
    \caption{Mean conditional weights $\{\epsilon_i\}$ within each class vs.\ training epochs on CIFAR-LT-10 (left: IF = 100; right: IF = 10).}
    \label{fig:eps_analysis}
\end{figure}

\vspace{-10pt}
\paragraph{What are the learned conditional weights?} We are interested in examining the conditional weights $\{\epsilon_i\}$ for each class throughout the training. For a visualization purpose, we average them within each class. Figure~\ref{fig:eps_analysis} demonstrates how they change over the last 20 epochs for the 1st, 4th, 7th, and 10th classes of CIFAR-LT-10. The two panels correspond to the imbalance factors of 100 and 10, respectively. Interestingly, the learned conditional weights of the tail classes are more prominent than those of the head classes in most epochs. Moreover, the conditional weights of the two head classes (the 1st and 4th) are even below 0 at certain epochs. Such results verify our intuition that the scarce training examples of the tail classes deserve more attention in training to make the neural network perform in a balanced fashion at the test phase.

\begin{table}
\centering
\caption {\label{tab:iNat-LT} Classification errors on iNat 2017 and 2018. (*results reported in paper. CE=cross-entropy, CB=class-balanced)}
\vspace{2pt}
\resizebox{0.45\textwidth}{!}{%
\begin{tabular}{l|c|c||c|c}
\hline

\hline

\hline
Dataset & \multicolumn{2}{c||}{iNat 2017} & \multicolumn{2}{c}{\scriptsize iNat 2018} \\ \hline
 Method & Top-1 & Top-3/5 & \scriptsize{Top-1} & {\scriptsize Top-3/5}  \\ \hline
CE & 43.49 & 26.60/21.00 & \scriptsize{36.20} & \scriptsize{19.40/15.85}   \\ \hline
CB CE~\cite{CBLoss} & 42.59
 & 25.92/20.60 & \scriptsize{34.69}  & \scriptsize{19.22/15.83} \\ \hline

{\bf Ours, CE} & {\bf 40.62} & \textbf{23.70/18.40} & \scriptsize{\bf 32.45} & \scriptsize{\textbf{18.02/13.83}}  \\ \hline

\hline

\hline

CB focal~\cite{CBLoss}* & 41.92 & --/20.92 
& \scriptsize{38.88}   & \scriptsize{--/18.97} \\ \hline 


LDAM~\cite{LDAM}* & -- & -- &  \scriptsize{35.42}  & \scriptsize{--/16.48} \\ \hline

LDAM-drw* & -- & --
&  \scriptsize{32.00}  & \scriptsize{--/14.82} \\ \hline



cRT~\cite{kang2019decoupling}* & -- & --
&  \scriptsize{34.8}  & \scriptsize{--} \\ \hline 
cRT+epochs* & -- & --
&  \scriptsize{32.4}  & \scriptsize{--} \\ \hline 


\end{tabular}
}%
\vspace{-12pt}
\end{table}

\begin{table}
\centering
\caption {\label{tab:Ablation_CIFAR-10} Ablation study of our approach by using the cross-entropy loss on CIFAR-LT-10. The results are test top-1 errors\%.}
\vspace{2pt}
\resizebox{0.45\textwidth}{!}
{%
\begin{tabular}{l|c|l|l}
\hline

Imbalance factor & 100 & \multicolumn{1}{c|}{50} & \multicolumn{1}{c}{20} \\ \hline


 
 

L2RW~\cite{L2RW} & 27.77  & 23.55  & 18.65 \\ \hline
 
L2RW, pre-training & 25.96  & 22.04  & 15.67 \\ \hline

L2RW, pre-training, init.\ by $w_{y}$ & 26.26  & 22.50  & 17.44 \\ \hline

L2RW, pre-training, $w_{y_i}+\epsilon_i$ & 24.54  & 20.47  & 14.38 \\ \hline

Ours  & \multicolumn{1}{l|}{\bf 23.59} & {\bf 19.49} & {\bf 13.54} \\ \hline

\hline

\hline

Ours updating $w_{y}$ & 25.42  & 20.13 & 15.62  \\ \hline

Class-balanced~\cite{CBLoss} & 27.63  & 21.95 & 15.64 \\ \hline
\end{tabular}
}%
\vspace{-15pt}
\end{table}

\vspace{-10pt}
\paragraph{Ablation study: ours vs.\ L2RW.}
Our overall algorithm differs from L2RW mainly in four ways: 1) pre-training the network, 2) initializing the weights by \textit{a priori} knowledge, 3) two-component weights, and estimating the class-wise components by a separate algorithm~\cite{CBLoss}, and 4) no clipping or normalization of the weights. Table~\ref{tab:Ablation_CIFAR-10} examines these components by applying them one after another to L2RW. First, pre-training the neural network boosts the performance of the vanilla L2RW. Second, if we initialize the sample weights by our class-wise weights $\{w_y\}$, the errors increase a little probably because the clipping and normalization steps in L2RW require more careful initialization to the sample weights. Third, if we replace the sample weights by our two-component weights, we can bring the performance of L2RW closer to ours. Finally, after we remove the clipping and normalization, we arrive at our algorithm, which gives rise to the best results among all variations. 

\vspace{-10pt}
\paragraph{Ablation study: the two-component weights.}
By Table~\ref{tab:Ablation_CIFAR-10}, we also highlight the importance of the two-component weights $\{w_{y_i}+\epsilon_i\}$ motivated from our domain adaptation point of view to the long-tailed visual recognition. First of all, they benefit L2RW (comparing ``L2RW, pre-training, $w_{y_i}+\epsilon_i$'' with ``L2RW, pre-training'' in Table~\ref{tab:Ablation_CIFAR-10}). Besides, they are also vital for our approach. If we drop the class-wise weights, our results would be about the same as L2RW with pre-training. If we drop the conditional weights and meta-learn the class-wise weights (cf.\ ``Ours updating $w_y$''), the errors become larger than our original algorithm. Nonetheless, the results are better than the class-balanced training (cf.\ last row in Table~\ref{tab:Ablation_CIFAR-10}), implying that the learned class-wise weights give rise to better models than the effective-number-based class-wise weights~\cite{CBLoss}. 


\subsection{Object recognition with iNat 2017 and 2018}
We use ResNet-50~\cite{Resnet} as the backbone network for the iNat 2017 and 2018 datasets. The networks are pre-trained on ImageNet for iNat 2017 and on ImageNet plus iNat 2017 for iNat 2018.  We experiment with the mini-batch size of 64 and the learning rate of 0.01. We train all the models using the stochastic gradient descent with momentum. For the meta-learning stage of our approach, we switch to a small learning rate, 0.001.

Table~\ref{tab:iNat-LT} shows the results of our two-component weighting applied to the cross-entropy loss. We shrink the text size for iNat 2018 to signify that we advocate experiments with iNat \textit{2017} instead  because there are only three validation/test images per class in iNat 2018 (cf.\  Table~\ref{tab:datasets}). Our approach boosts the cross-entropy training by about 2\% more than the class-balanced weighting~\cite{CBLoss} does. As we have reported similar effects for the focal loss and the LDAM loss on CIFAR-LT with extensive experiments, we do not run them on the large-scale iNat datasets to save computation costs. Nonetheless, we include the results reported in the literature of the focal loss, LDAM loss, and a classifier re-training method~\cite{kang2019decoupling}, which was published after we submitted the work to CVPR 2020.

\begin{table}
\centering

\caption {\label{tab:largescale-LT} Classification errors on ImageNet-LT and Places-LT. (*reported in paper. CE=cross-entropy, CB=class-balanced)}
\vspace{2pt}
\resizebox{0.45\textwidth}{!}{%
\begin{tabular}{l|c|c||c|c}
\hline
Dataset & \multicolumn{2}{c||}{ImageNet-LT} & \multicolumn{2}{c}{Places-LT} \\ \hline
Method & Top-1 & Top-3/5 & Top-1 & Top-3/5 \\ \hline
CE & 74.74 &  61.35/52.12 & 73.00 & 52.05/41.44   \\ \hline
CB CE~\cite{CBLoss} & 73.41 & 59.22/50.49  & 71.14  & 51.58/41.96 \\ \hline
{\bf Ours, CE} & {\bf 70.10} & {\bf 53.29/45.18}
 & {\bf 69.20} & \textbf{47.95/38.00} \\ \hline
 






\end{tabular}
}%
\vspace{-5pt}
\end{table}

\subsection{Experiments with ImageNet-LT and Places-LT}
Following Liu et al.'s experiment setup~\cite{OLTR}, we employ ResNet-32 and ResNet-152 for the experiments on ImageNet-LT and Places-LT, respectively. For ImageNet-LT, we adopt an initial learning rate of 0.1 and decay it by 0.1 after every 35 epochs. For Places-LT, the initial learning rate is 0.01 and is decayed by 0.1 every 10 epochs. For our own approach, we switch from the cross-entropy training to the meta-learning stage when the first decay of the learning rate happens. The mini-batch size is 64, and the optimizer is stochastic gradient descent with momentum. 

\vspace{-10pt}
\paragraph{Results.}
Table~\ref{tab:largescale-LT} shows that the class-balanced training improves the vanilla cross-entropy results, and our two-component weighting further boosts the results. We expect the same observation with the focal and LDAM losses. Finally, we find another improvement by updating the classification layers only in the meta-learning stage. We arrive at 62.90\% top-1 error (39.86/29.87\% top-3/5 error) on Places-LT, which is on par with 64.1\% by OLTR~\cite{OLTR} or 63.3\% by cRT~\cite{kang2019decoupling}, while noting that our two-component weighting can be conveniently applied to both OLTR and cRT.

\section{Conclusion}
In this paper, we make two major contributions to the long-tailed visual recognition. One is the novel domain adaptation perspective for analyzing the mismatch problem in long-tailed classification. While the training set of real-world objects is often long-tailed with a few classes that dominate, we expect the learned classifier to perform equally well in all classes. By decomposing this mismatch into class-wise differences and the discrepancy between class-conditioned distributions, we uncover the implicit assumption behind existing class-balanced methods, that the training and test sets share the same class-conditioned distribution. Our second contribution is to relax this assumption to explicitly model the ratio between two class-conditioned distributions. Experiments on six datasets verify the effectiveness of our approach.

\vspace{-10pt}
\paragraph{Future work.} We shall explore other techniques~\cite{bickel2009discriminative,sugiyama2007covariate} for estimating the conditional weights. In addition to the weighting scheme, other domain adaptation methods~\cite{gopalan2015domain,csurka2017domain}, such as learning domain-invariant features~\cite{ganin2014unsupervised,tzeng2017adversarial} and data sampling strategies~\cite{gong2013connecting,rai2010domain}, may also benefit the long-tailed visual recognition problem. Especially, the domain-invariant features align well with Kang et al.'s recent work on decoupling representations and classifications for long-tailed classification~\cite{kang2019decoupling}. 

\vspace{-10pt}
\paragraph{Acknowledgements.}The authors  thank the support of NSF awards 1149783, 1741431, 1836881, and 1835539.  

{\small
\bibliographystyle{ieee_fullname}
\bibliography{references}
}


\newpage
\appendix
\section*{Appendices}
\addcontentsline{toc}{section}{Appendices}
\renewcommand{\thesubsection}{\Alph{subsection}}

\begin{itemize}
\item{Multiple runs of experiments on CIFAR-LT-10 under different imbalance factors (IFs) (Section~\ref{sec:multiple}).}
\item{Detailed comparison of various methods on large-scale long-tailed datasets (Section~\ref{sec:study}).}
\end{itemize}

\section{Multiple runs on CIFAR-LT-10} \label{sec:multiple}
In this experiment, we further validate our approach by running each setting 5 times with different random seeds. Table~\ref{tab:multipleruns} shows the mean top-1 errors (\%) and the standard deviations under the imbalance factors of 200, 100, and 50. We can see that the mean error rates are consistent with the results provided in Table 2 of the main paper.

\begin{table*}
\centering
\caption {\label{tab:multipleruns} Multiple runs of our approach by using the cross-entropy loss on CIFAR-LT-10. The results are top-1 errors\% on the test sets. }
\vspace{5pt}


\begin{tabular}{l|c|l|l}
\hline

\hline

\hline
Imbalance factor & 200 & \multicolumn{1}{c|}{100} & \multicolumn{1}{c}{50} \\ \hline

Cross-entropy training & 34.32 & 29.64 & 25.19  \\ \hline

Class-balanced cross-entropy loss~\cite{CBLoss} & 31.11 & 27.63 & 21.95 \\ \hline

\begin{tabular}[c]{@{}l@{}}Class-balanced fine-tuning~\cite{Cui-Finetune}\\ Class-balanced fine-tuning~\cite{Cui-Finetune}*\end{tabular} & \begin{tabular}[c]{@{}c@{}}33.76\\ 33.92\end{tabular} & \begin{tabular}[c]{@{}l@{}}28.66\\ 28.67\end{tabular} & \begin{tabular}[c]{@{}l@{}}22.56\\ 22.58\end{tabular}  \\ \hline

\begin{tabular}[c]{@{}l@{}}L2RW~\cite{L2RW}\\ L2RW~\cite{L2RW}*\end{tabular} & \begin{tabular}[c]{@{}c@{}}33.75\\ 33.49\end{tabular} & \begin{tabular}[c]{@{}l@{}}27.77\\ 25.84\end{tabular} & \begin{tabular}[c]{@{}l@{}}23.55\\ 21.07\end{tabular}  \\ \hline

Meta-weight net~\cite{MWN} & 32.8 & 26.43 & 20.9  \\ \hline

Ours with cross-entropy loss & \textbf{29.32 $\pm$ 0.23}  & \textbf{23.71 $\pm$ 0.22}  & \textbf{19.45 $\pm$ 0.28} \\ \hline
 
\end{tabular}

\end{table*}










\section{Detailed comparison of various methods on large-scale long-tailed datasets} \label{sec:study}
In this section, we present the top-1 errors (\%) of various methods for ImageNet-LT, Places-LT, and iNaturalist 2018\footnote{While we include the comparison on iNaturalist 2018 due to that most existing related works report results on this dataset, we reiterate that we advocate the use of iNaturalist 2017, instead of 2018, in this and future work due to the extremely small validation set of iNaturalist 2018.}. As the experiment setups of the existing works vary by network initialization, the sampling strategy of mini-batches, losses, trainable layers of a network, etc., it is hard to have a fair comparison by the end results. Hence, besides their top-1 errors, we also report the experiment setups for each method. Tables~\ref{tab:ImageNet-LT},~\ref{tab:Places-LT},~and~\ref{tab:iNat18} show the results of the different methods on ImageNet-LT, Places-LT, and iNaturalist 2018, respectively. Our approach outperforms the class-balanced weighting scheme for both the cross-entropy loss and the focal loss, as we observed in the main paper. Moreover, our results are on par with the best reported ones except on ImageNet-LT. Finally, we stress that almost all existing methods employ a class-balanced weighting or sampling strategy no matter what their main techniques are to tackle the long-tailed problem. Hence, given our consistent improvements over the class-balanced weighting, we expect the methods which have benefited from the class-balancing can gain further from our two-component weighting.


\begin{table*}[h!]
\centering
\caption {\label{tab:ImageNet-LT} Test top-1 errors (\%) of different methods on ImageNet-LT. * indicates the re-run results.}
\vspace{2pt}
\resizebox{1.0\textwidth}{!}{%
\begin{tabular}{l|l|l|l|l|l|l|l}
\hline

\hline

\hline
Methods & \multicolumn{1}{c|}{NN} & Initialization & Sampling & Loss & \begin{tabular}[c]{@{}l@{}}Stage-1\\ Trainable Variables\end{tabular} & \begin{tabular}[c]{@{}l@{}}Stage-2\\ Trainable Variables\end{tabular} & Results \\ \hline

Vanilla Model & ResNet-10 & No-pretrain & Class-Balanced &  CE & All  & - & 80.0 \\ \hline

Vanilla Model~\cite{Focalloss} & ResNet-10 &No-pretrain  & Class-Balanced & Focal &All   &-  & 69.8 \\ \hline

Vanilla Model & ResNet-10 & No-pretrain  &Class-Balanced  &  Lifted&All  &-  & 69.2 \\ \hline

Vanilla Model~\cite{Rangeloss} & ResNet-10 & No-pretrain & Class-Balanced &  Range & All & - & 69.3 \\ \hline

Joint~\cite{kang2019decoupling} &ResNet-10  & No-pretrain &Class-Balanced  & CE & All & All & 65.2 \\ \hline

NCM~\cite{kang2019decoupling} & ResNet-10 & No-pretrain &Class-Balanced  & CE &All  & Classifier layer & 64.5  \\ \hline

cRT~\cite{kang2019decoupling} & ResNet-10 & No-pretrain &Class-Balanced  &CE  &All  &  Classifier layer& \textbf{58.2} \\ \hline

$\tau$-normalized~\cite{kang2019decoupling} & ResNet-10 & No-pretrain &Class-Balanced  &  CE&All  &Classifier layer  & \underline{59.4}    \\ \hline

OLTR*~\cite{OLTR} & ResNet-10  & No-pretrain &Class-Balanced  &CE  &All  & All & 65.6  \\ \hline

OLTR~\cite{OLTR} & ResNet-10 & No-pretrain &Class-Balanced  &CE  &All  & All & 64.4 \\ \hline

Ours & ResNet-10 & No-pretrain & None  &CE  & All & Classifier layer & 63.5 \\ \hline

Ours & ResNet-10  & No-pretrain & None & Focal &All  &Classifier layer  & 63.3 \\ \hline

\hline

\hline

Vanilla Model & ResNet-50  & No-pretrain & None & CE & All  & -  & 59.0 \\ \hline

CB~\cite{CBLoss} & ResNet-50 & No-pretrain & None &  CE& All & - & 58.2 \\ \hline
Joint~\cite{kang2019decoupling} &ResNet-50  & No-pretrain &Class-Balanced  & CE & All & All & 58.4 \\ \hline

NCM~\cite{kang2019decoupling} & ResNet-50 & No-pretrain &Class-Balanced  & CE &All  & Classifier layer & 55.7  \\ \hline

cRT~\cite{kang2019decoupling} & ResNet-50 & No-pretrain &Class-Balanced  &CE  &All  &  Classifier layer& \underline{52.7} \\ \hline

$\tau$-normalized~\cite{kang2019decoupling} & ResNet-50 & No-pretrain &Class-Balanced  &  CE&All  &Classifier layer  & 53.3   \\ \hline

Ours & ResNet-50  & No-pretrain & None & CE &All  &Classifier layer  &  \textbf{52.0} \\ \hline

\end{tabular}
}%
\end{table*}


\begin{table*}[h!]
\centering
\caption {\label{tab:Places-LT} Test top-1 errors (\%) of different methods on Places-LT. * indicates the re-run results.}
\vspace{2pt}
\resizebox{1.0\textwidth}{!}{%
\begin{tabular}{l|l|l|l|l|l|l|l}
\hline

\hline

\hline
Methods & \multicolumn{1}{c|}{NN} & Initialization & Sampling & Loss & \begin{tabular}[c]{@{}l@{}}Stage-1\\ Trainable Variables\end{tabular} & \begin{tabular}[c]{@{}l@{}}Stage-2\\ Trainable Variables\end{tabular} & Results \\ \hline
Vanilla Model & ResNet-152  & ImageNet & Class-Balanced  & CE & \begin{tabular}[c]{@{}l@{}} FC layers\\ Last Block + FC\end{tabular} & \begin{tabular}[c]{@{}l@{}}-\\ -\end{tabular} & \begin{tabular}[c]{@{}l@{}} 72.1\\ 69.7\end{tabular} \\ \hline
Vanilla Model~\cite{Focalloss} & ResNet-152  & ImageNet & Class-Balanced  & Focal & \begin{tabular}[c]{@{}l@{}} FC layers\\ Last Block + FC\end{tabular} & \begin{tabular}[c]{@{}l@{}}-\\ -\end{tabular} & \begin{tabular}[c]{@{}l@{}}67.0\\ 66.5\end{tabular} \\ \hline
Vanilla Model & ResNet-152  & ImageNet & Class-Balanced & Lifted & FC layers & - & 64.8 \\ \hline
Vanilla Model~\cite{Rangeloss} & ResNet-152 &ImageNet  &Class-Balanced  & Range&  FC layers & - & 64.9 \\ \hline

Joint~\cite{kang2019decoupling} & ResNet-152  &ImageNet  & Class-Balanced & CE & Last block + FC & Last block + FC & 69.8 \\ \hline

NCM~\cite{kang2019decoupling} &ResNet-152  &ImageNet  &Class-Balanced  & CE & Last block + FC  & Classifier layer & 63.7 \\ \hline

cRT~\cite{kang2019decoupling} &ResNet-152  &ImageNet  & Class-Balanced &CE  &Last block + FC  &Classifier layer  & 63.3 \\ \hline

$\tau$-normalized~\cite{kang2019decoupling} & ResNet-152 & ImageNet &Class-Balanced  & CE &Last block + FC  & Classifier layer  &\textbf{62.1}  \\ \hline

OLTR*~\cite{OLTR} &ResNet-152  &ImageNet  & Class-Balanced &CE  &Last block + FC  & FC + memory  & 64.8 \\ \hline

OLTR~\cite{OLTR} & ResNet-152 &ImageNet  &Class-Balanced  & CE & Last block + FC & FC + memory & 64.1 \\ \hline

Ours &ResNet-152  &ImageNet  & None  & CE &Last block + FC  & Classifier layer & 62.9 \\ \hline

Ours & ResNet-152 &ImageNet  & None & Focal &Last block + FC  & Classifier layer & \underline{62.2} \\ \hline

\end{tabular}
}%
\end{table*}


\begin{table*}[h!]
\centering
\caption {\label{tab:iNat18} Test top-1 errors (\%) of different methods on iNaturalist 2018.}
\vspace{2pt}
\resizebox{1.0\textwidth}{!}{%

\begin{tabular}{l|l|l|l|l|l|l|l}
\hline

\hline

\hline
Methods & \multicolumn{1}{c|}{NN} & Initialization & Sampling & Loss & \begin{tabular}[c]{@{}l@{}}Stage-1\\ Trainable Variables\end{tabular} & \begin{tabular}[c]{@{}l@{}}Stage-2\\ Trainable Variables\end{tabular} & Results \\ \hline

Vanilla Model & ResNet-50 & No-pretrain & None & CE & All & -& 42.9 \\ \hline
Vanilla Model & ResNet-50  & ImageNet+iNat'17  & None  & CE  & All & - & 36.2 \\ \hline

LDAM~\cite{LDAM} & ResNet-50 & No-pretrain  & None & LDAM & All & - & 35.4 \\ \hline
LDAM-DRW~\cite{LDAM} & ResNet-50  & No-pretrain  & None & LDAM & All &-  & \textbf{32.0} \\ \hline

CB~\cite{CBLoss} & ResNet-50 & ImageNet+iNat'17 & None &  CE& All & - & 34.7 \\ \hline
CB~\cite{CBLoss} & ResNet-50 & No-pretrain &None  &Focal  & All & - & 38.9 \\ \hline

Joint~\cite{kang2019decoupling} & ResNet-50 &No-pretrain  &Class-Balanced  & CE & All & All  & 38.3 \\ \hline

NCM~\cite{kang2019decoupling} &ResNet-50  & No-pretrain & Class-Balanced &CE  &All  & Classifier layer  & 41.8 \\ \hline

cRT~\cite{kang2019decoupling} &ResNet-50  & No-pretrain &Class-Balanced  & CE & All & Classifier layer & 34.8 \\ \hline

$\tau$-normalized~\cite{kang2019decoupling} &ResNet-50  & No-pretrain & Class-Balanced & CE & All & Classifier layer & 34.4 \\ \hline

Ours &ResNet-50  & ImageNet+iNat'17 & None &CE  & All & All & \underline{32.4} \\ \hline

Ours &ResNet-50  & ImageNet+iNat'17 & None &Focal  & All & All & \underline{32.3} \\ \hline

\hline

\hline

Vanilla Model & ResNet-101  & ImageNet+iNat'17  & None  & CE  & All & - & 34.3 \\ \hline

CB~\cite{CBLoss} & ResNet-101  & ImageNet+iNat'17  & None  & CE  & All & - & 32.7 \\ \hline

Ours &ResNet-101  & ImageNet+iNat'17 & None & CE  & All & All & \textbf{31.5} \\ \hline
\end{tabular}
}%
\end{table*}

\end{document}